\documentclass[twocolumn]{article}
\usepackage{spconf,amsmath,graphicx}

\usepackage{enumitem}
\setlist{nosep, leftmargin=*}
\usepackage[table]{xcolor}
\usepackage{amssymb, amsfonts}
\usepackage{euscript}
\usepackage{mathtools}
\usepackage{booktabs}
\usepackage{multirow}
\usepackage[colorlinks=true, linkcolor=blue, citecolor=green, urlcolor=red]{hyperref}

\usepackage{graphicx}
\usepackage{subcaption}
\usepackage{cleveref}
\graphicspath{Figures/}

\title{HCA-Net: Hierarchical Context Attention Network for Intervertebral Disc Semantic Labeling}

\name{Afshin Bozorgpour $^{1}$\!\! \quad Bobby Azad $^2$\!\! \quad Reza Azad $^{3}$\!\! \quad Yury Velichko $^{4}$\!\! \quad Ulas Bagci $^{4}$\!\! \quad Dorit Merhof $^{1,5}$}

 \address{
 $^{1}$Faculty of Informatics and Data Science, University of Regensburg, Germany \\
 $^2$South Dakota State University, Brookings, USA \\ 
 $^{3}$Faculty of Electrical Engineering and Information Technology, RWTH Aachen University, Germany \\ 
 $^{4}$Machine and Hybrid Intelligence Lab, Northwestern University, Chicago, IL, USA  \\
 $^{5}$Fraunhofer Institute for Digital Medicine MEVIS, Germany }

\begin{document}
%
\maketitle
\begin{abstract}
Accurate and automated segmentation of intervertebral discs (IVDs) in medical images is crucial for assessing spine-related disorders, such as osteoporosis, vertebral fractures, or IVD herniation. We present HCA-Net, a novel contextual attention network architecture for semantic labeling of IVDs, with a special focus on exploiting prior geometric information. Our approach excels at processing features across different scales and effectively consolidating them to capture the intricate spatial relationships within the spinal cord.
To achieve this, HCA-Net models IVD labeling as a pose estimation problem, aiming to minimize the discrepancy between each predicted IVD location and its corresponding actual joint location. In addition, we introduce a skeletal loss term to reinforce the model's geometric dependence on the spine. This loss function is designed to constrain the model's predictions to a range that matches the general structure of the human vertebral skeleton. As a result, the network learns to reduce the occurrence of false predictions and adaptively improves the accuracy of IVD location estimation.
Through extensive experimental evaluation on multi-center spine datasets, our approach consistently outperforms previous state-of-the-art methods on both MRI T1w and T2w modalities. The codebase is accessible to the public on \href{https://github.com/xmindflow/HCA-Net}{GitHub}.
\end{abstract}
\section{Introduction}
\label{sec:intro}
The human spinal column comprises 33 individual vertebrae, organized in a stacked configuration and interconnected by ligaments and intervertebral discs (IVDs), commonly referred to as IVDs. This anatomical structure is further categorized into five distinct regions, including the cervical, thoracic, lumbar, sacral, and caudal vertebrae~\cite{gewiess2023influence}. Each of these regions plays a critical role in various physiological functions, such as shock absorption, load bearing, spinal cord protection, and load distribution management~\cite{al2022transfer}.
IVDs are fibrocartilaginous cushions that serve as primary articulations between adjacent vertebrae. They play a critical role in absorbing the forces and shocks exerted on the body during movement, ensuring spinal flexibility while preventing vertebral friction. Any disruption to the structural integrity of IVDs, whether due to aging, degeneration, or injury, can alter the properties and affect the mechanical performance of the surrounding tissues. Consequently, the precise localization and segmentation of IVDs are critical steps in the diagnosis of spinal disorders and provide invaluable insights into the efficacy of treatment modalities.
To address this challenge, numerous semi-automated and fully automated methods have been proposed in the literature~\cite{adibatti2023segmentation,rouhier2020spine,azad2022intervertebral,hou2023mri}.

Gros et al.~\cite{gros2018automatic} proposed a local descriptor-based method to detect the C2/C3 intervertebral disc (IVD) in medical imaging. This technique compares the mutual information between a patient's image and a template to find the region closest to the spine template. This handcrafted approach generally yields good results, but its performance degrades significantly when the patient's images deviate significantly from the template. To overcome these limitations of manual methods, deep learning models have been employed for robust IVD labeling.
Chen et al.~\cite{chen2019vertebrae} introduced a 3D CNN model for MRI data to enabling 3D segmentation and accurate identification of vertebral disc locations. Cai et al.~\cite{cai2015multi} utilized a 3D Deformable Hierarchical Model for 3D spatial vertebral disc localization. Rouhier et al.~\cite{rouhier2020spine} trained a Count-ception model on 2D MRI sagittal slices to detect vertebral discs.
Adibatti et al.~\cite{adibatti2023segmentation} proposed a capsule stacked autoencoder for IVD segmentation. Vania et al.~\cite{vania2021intervertebral} introduced a multi-optimization training system at various stages to enhance computational efficiency, building upon Mask R-CNN. Meanwhile, Wimmer et al.~\cite{wimmer2018fully} presented a cross-modality method for detecting both vertebral and intervertebral discs in volumetric data, using a local entropy-based texture model followed by alignment and refinement techniques.
Mbarki et al.~\cite{mbarki2020lumbar} employed transfer learning to detect lumbar discs from axial images using a 2D convolutional structure. Their network, based on the U-Net structure with a VGG backbone, generated a spine segmentation mask used to calculate herniation in lumbar discs.
Azad et al.~\cite{azad2021stacked} redefined semantic vertebral disc labeling as pose estimation by implementing an hourglass neural network for semantic labeling of IVDs. In a more recent approach~\cite{azad2022intervertebral}, they propose an enhancement to the detection process by including the image gradient as an auxiliary input to better capture and represent global shape information.

\begin{figure}[!b]
\includegraphics[width=\columnwidth]{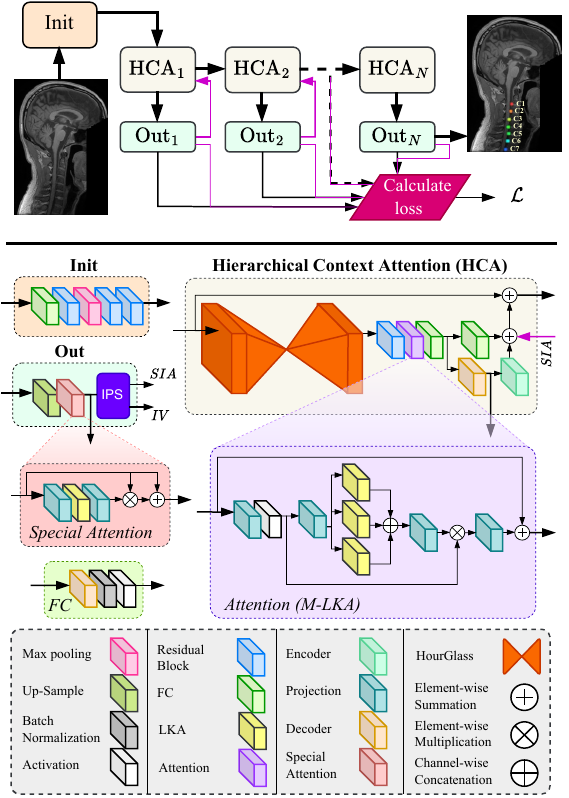}
\caption{Structure of the proposed HCA-Net method for IVD semantic labeling.} \label{fig:proposed_method}
\end{figure}

Existing methods have attempted  to improve shape information by incorporating image gradients as auxiliary data~\cite{azad2022intervertebral}, focusing on vertebral column region detection~\cite{vania2021intervertebral}, and modeling pose information~\cite{azad2021stacked}. However, these methods still face limitations in implicitly conditioning the representation space using global vertebral column information to efficiently model geometric constraints. As a result, these strategies may lead to undesirable false positive and false negative predictions. To address this challenge, we present HCA-Net, a novel pose estimation approach that leverages a robust framework featuring Multi-scale Large Kernel Attention (M-LKA) modules to facilitate the comprehensive capture of contextual information while preserving local intricacies. This architectural enhancement plays a pivotal role in enabling precise semantic labeling.
Furthermore, to enhance the model's reliance on vertebral column geometry, we introduce the skeleton loss function to effectively constrains the model's predictions within a range consistent with the human vertebral skeleton.
Our key contributions are: \textbf{(1)} A contextual attention network for semantic labeling, which incorporates the multi-scale large kernel attention mechanism to model both local and global representations, \textbf{(2)} the skeleton loss function to implicitly enforce geometrical information of the vertebral column into the model prediction.

\section{Method}
The design of our contextual attention network for IVD labeling is driven by the need to extract information from medical images at different scales. While local features are essential for discerning specific anatomical structures such as IVDs, achieving precise disc labeling requires a holistic understanding of the entire spinal structure. This includes considerations such as the orientation of the spine, the arrangement of the IVDs, and the relationships between neighboring discs, which are most effectively captured at different scales within the medical image. To address this challenge, we introduce our novel hierarchical context attention strategy, illustrated in \Cref{fig:proposed_method}. Our approach incorporates multi-scale, large kernel attention blocks to capture both local and global dependencies, while constraining the model prediction with prior information on the distribution of the IVDs.

\subsection{Network Architecture}
The architecture of the HCA-Net is structured as follows: First, a sequence of convolutional layers is applied to process the input MRI image and transform it into a latent representation. Next, a hierarchical context attention module is employed to capture multi-scale representations. This module uses an hourglass block~\cite{newell2016stacked} to effectively model local representations, and then leverages large kernel attention across multiple scales to adjust the representation space based on local-to-global information, facilitating the incorporation of both local and long-range dependencies.

 \Cref{fig:proposed_method} illustrates the construction of HCA-Net, which involves stacking hierarchical context attention (HCA) blocks and incorporates the process of learning object pose estimation through ($N$-1) intermediate predictions $\text{Out}_j$ along with one final prediction. This approach takes into account the multilevel representations generated by the $N$-stacked HCA blocks. Finally, we merge the intermediate and final prediction masks using the $1\times1$ convolution, resulting in a $V$ channel prediction map ($\hat{y}$). Each channel within this map corresponds to a specific intervertebral location, thus providing a comprehensive representation of intervertebral positions.
To minimize the network's prediction error, we take the sum of mean squared error (MSE) loss between the network prediction $\hat{y}$ and the ground truth $y$:

\begin{equation}
\mathcal{L}_v=\frac{1}{V \times M} \sum_{i=1}^{V}\sum_{p=1}^{M}\left(y^i_p-\hat{y}^i_p\right)^{2},
\label{eq:trainingloss}
\end{equation}
where $M$ corresponds to the number of pixels in the ground truth mask.
To reinforce the incorporation of vertebral column structure as an additional supervisory signal to enhance network predictions, we introduce the ``skeleton loss" $\mathcal{L}_{sk}$ term to the overall loss function. Consequently, during each training step, HCA-Net aims to minimize the combined loss function:
\begin{equation}
\label{eq:combo}
\mathcal{L} = \mathcal{L}_{v}+ \lambda \mathcal{L}_{sk}
\end{equation}

\subsubsection{Multi-scale Large Kernel Attention (M-LKA)}
Achieving accurate semantic labeling of IVDs requires the consideration of both local and global semantic representations. Given the geometrical interdependencies among intervertebral joint locations, relying solely on local representations may result in erroneous predictions. To overcome these challenges, we introduce an innovative approach that leverages the Large Kernel Attention (LKA) mechanism. We enhance the LKA module by extending it across multiple scales. The rationale behind this enhancement is to efficiently capture and integrate information at various spatial resolutions, which is especially valuable for tasks demanding precise predictions. In contrast to the original LKA, which employs fixed-sized filters and faces challenges in fully capturing information at different scales within an image, our M-LKA module utilizes parallel filters of varying sizes. This approach allows us to capture both fine-grained details and high-level semantic information concurrently.

The LKA module decomposes a $C \times C$ convolution into three components: a $[\frac{c}{d}]\times [\frac{c}{d}]$ depth-wise dilation convolution ($DW\text{-}D\text{-}Conv$) for long-range spatial convolution, a $(2d-1)\times(2d-1)$ depth-wise convolution ($DW\text{-}Conv$) for local spatial convolution, and a $1\times1$ convolution for channel-wise convolution. This decomposition enables us to extract long-range relationships within the feature space while maintaining computational efficiency and a manageable parameter count when generating the attention map. We further extend the LKA module into multiscale form as follows:

Let $\operatorname{F_S}(x)$ represent a set of feature maps obtained by applying depth-wise convolution (\text{DW-Conv}) to the input features $F(x)$ for each scale $s \in \mathbb{S}$. Then, $\operatorname{F_S}(x)$ can be expressed as:
$$
\operatorname{F_S}(x)=\left\{(\text{DW-Conv}(F(x)))_s \mid s \in \mathbb{S}\right\}
$$

Subsequently, the attention map $\text{Attention}$ is generated by applying a $1 \times 1$ convolution ($\operatorname{Conv}_{1 \times 1}$) to the feature maps obtained through depth-wise dilation convolution (\text{DW-D-Conv}) of $\operatorname{F_s}(x)$:
$$
\text{Attention}=\operatorname{Conv}_{1 \times 1}(\text{DW-D-Conv}(\operatorname{F_S}(x)))
$$

Finally, the output $\text{x'}$ is computed as the element-wise multiplication ($\otimes$) between the attention map $\text{Attention}$ and the input features $F(x)$:
$$
\text{x'}=\text{Attention} \otimes F(x)
$$

\subsection{Skeleton Loss Function}
Accurate IVD semantic labeling often faces the challenge of generating false predictions, necessitating a mechanism for guiding the network towards more reliable outcomes. To tackle this issue, we leverage the network's prediction map, denoted as $\hat{y}$, and apply the softmax operation to transform it into a 2D positional probability distribution for the IVD location in each channel:

\begin{equation}
    \mathbf{P}^{i}_j=\frac{\sigma({\hat{y}^i})}{\sum_p^M\sigma(\hat{y}^i)_p},
    \label{eq:pmf_joint}
\end{equation}

where $\mathbf{P}_j^i$ represents the probability of the respective intervertebral joint location within each channel.
Subsequently, we use the probability map to generate prototypes for each intervertebral location through $T$ times sampling from each channel and averaging as follows:

\[
\mathbf{V}^i_j = \frac{1}{T} \sum_{T} \text{Sampler}(\mathbf{P}^{i}_j),
\]

The $sampler$ function utilizes the probability map $\mathbf{P}^{i}_j$ to extract intervertebral locations in each channel. Subsequently, our approach integrates a distance function denoted as $D: \mathbb{R}^M \times \mathbb{R}^M \rightarrow [0,+\infty)$ to minimize the distance between the intervertebral column and the ground truth location. To this end, we model the skeleton loss function as follows: 
\begin{equation*}
    \begin{aligned}
        &\mathcal{L}_{sk} = \sum_{j=1}^N \left(\beta \mathcal{L}_j^{id}+(1-\beta)\mathcal{L}_j^{pd}\right)
        \\
        &\mathcal{L}^{id}_j = ||\mathbf{V}^i_j - \mathbf{V}^{\text{GT}}||, \quad \mathcal{L}^{pd}_j = \text{PD}(\mathbf{V}^i_j, \mathbf{V}^{\text{GT}})
        \\
        &\text{PD}(\mathbf{V}, \mathbf{V}^{\text{GT}})=\sum^{C-1}_{c}\sum^{C}_{k=c}\alpha^{k-c}\left(D(\mathbf{V},c,k)-D(\mathbf{V}^{\text{GT}},c,k)\right)^2
    \end{aligned}
    \label{eq:loss-gen}
\end{equation*}

Here, we define the distance function as $\text{D}(V, i, k) = ||V_i - V_{i+k}||$. The parameter $\alpha$ represents a learnable weight, while $\mathcal{L}^{id}$ denotes the $L2$ distance between the vertebral column prototype and the ground truth. Additionally, $\mathcal{L}^{pd}$ quantifies the pair-wise distance (PD), ensuring the preservation of the geometrical relationships within the intervertebral skeleton structure.

\begin{table*} 
    \caption{Intervertebral disc semantic labeling on the spine generic public dataset. Note that \textbf{DTT} indicates Distance to target}
    \footnotesize
    \label{tab:ex}
    \centering
    \begin{tabular}{c||c||c}
        \hline
        \begin{tabular}{c}
            \multicolumn{1}{c}{\textbf{Method}} \\
            \hline
            \textbf{}\\
            \hline
            Template Matching~\cite{ullmann2014automatic}\\
            Countception~\cite{rouhier2020spine}\\
            Pose Estimation~\cite{azad2021stacked}\\
            Look Once\\
            \hline
            \rowcolor[rgb]{1,0.95,0.88}
            \textbf{HCA-Net without $\mathcal{L}_{sk}$}\\
            \rowcolor[rgb]{.9,0.95,1}
            \textbf{HCA-Net}\\
        \end{tabular} &
        \begin{tabular}{ccc}
            \multicolumn{3}{c}{\textbf{T1}}\\
            \hline
            \textbf{DTT (mm)} & \textbf{FNR (\%)} & \textbf{FPR (\%)}\\
            \hline
            1.97(±4.08) & 8.1 & 2.53  \\
            \textbf{1.03(±2.81)} & 4.24 & 0.9\\
            1.32(±1.33) & 0.32 & 0.0\\
            1.2(±1.90) & 0.7 & 0.0\\
            \hline
            \rowcolor[rgb]{1,0.95,0.88}
            1.27(±1.78) & 0.6 & 0.0\\
            \rowcolor[rgb]{.9,0.95,1}
            1.19(±1.08) & \textbf{0.3} & \textbf{0.0}\\
        \end{tabular} &
        \begin{tabular}{ccc}
            \multicolumn{3}{c}{\textbf{T2}}\\
            \hline
            \textbf{DTT (mm)} & \textbf{FNR (\%)} & \textbf{FPR (\%)}\\
            \hline
            2.05(±3.21) & 11.1 & 2.11\\
            1.78(±2.64) & 3.88 & 1.5\\
            1.31(±2.79) & 1.2 & 0.6\\
            1.28(±2.61) & 0.9 & 0.0\\
            \hline
            \rowcolor[rgb]{1,0.95,0.88}
            1.34(±2.28) & 1.2 & 0.0\\
            \rowcolor[rgb]{.9,0.95,1}
            \textbf{1.26(±2.16)} & \textbf{0.61} & \textbf{0.0}\\
        \end{tabular}
    \end{tabular}
\end{table*}

\section{Experimental Setup and Results}
\noindent\textbf{Experimental Setup:}
In our experiment, we use the Spine Generic Dataset~\cite{spinedataset} for IVD labeling. This dataset contains samples from 42 medical centers around the world in both T1-weighted (T1w) and T2-weighted (T2w) contrasts and exhibits a large variation in terms of quality, scale, and imaging device. To prepare the dataset for the training, we first calculate the average of six sagittal slices, centered on the middle slice, to create a representative data sample for each subject. To ensure uniformity and to minimize the impact of data variations, we normalize each image to the [0, 1] range. Next, using the IVD coordinate on the 2D position, we create a heatmap image by applying a Gaussian kernel convolution on each position of the IVD. Similar to~\cite{rouhier2020spine} we extract 11 IVDs for each subject. In instances where an IVD is missing, we designate its position as ``unknown" and mitigate its influence on the training process by effectively filtering it out using the visibility flag within the loss function.
Following~\cite{azad2021stacked}, we train the model for 500 epochs with RMSprob optimization using a learning rate of $2.5e-4$ and a batch size of 4. Our experimental hyperparameter settings entail $\lambda=2e-4$ (in Equation \ref{eq:combo}), $\beta=0.75$ (in Equation \ref{eq:loss-gen}) and $\alpha=0.8$ in the PD function. 
We follow evaluation metrics from prior studies~\cite{azad2021stacked, rouhier2020spine}, including L2 distance for predicted vs. ground truth IVD positions in 3D space. Additionally, we report False Positive Rate (FPR) and False Negative Rate (FNR).

\noindent\textbf{Results:}
\Cref{tab:ex} presents a comprehensive analysis of our HCA-Net compared to other SOTA methods for IVD semantic labeling. Our approach consistently outperforms existing methods in both T1w and T2w MRI modalities, showcasing its superior accuracy and reliability.
In T1w MRI, our method excels with an impressive average distance to the target (DTT) of $1.19$ mm, significantly outperforming other methods. This low DTT, combined with a standard deviation of only $1.08$ mm, makes our approach highly reliable for precise IVD localization. Notably, even without the $\mathcal{L}_{sk}$ module, our HCA-Net performs remarkably well, achieving a DTT of $1.27$ mm and displaying superior accuracy compared to the alternatives.
In the T2w MRI, our HCA-Net again enhances the performance, with an outstanding DTT of $1.26$ mm. This result significantly outperforms previous work, underlining the robustness and accuracy of our approach. Additionally, our method achieves a lower false negative rate (FNR) of $0.61\%$ in T2w, indicating its ability to capture IVDs effectively and minimize missed detections.

\begin{figure}[!ht]
\includegraphics[width=0.5\textwidth]{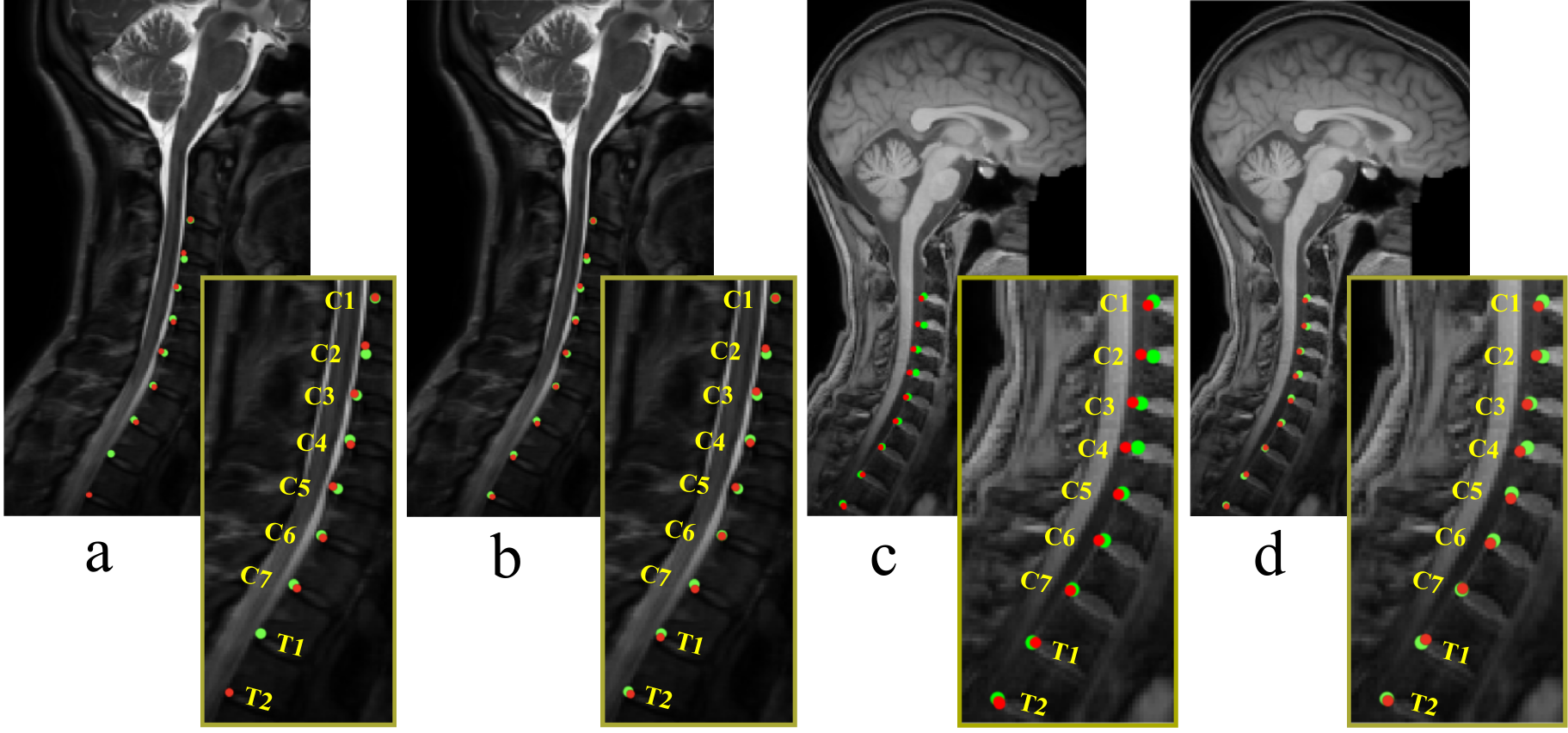}
\caption{Comparison of results on T1w (a-b) and T2w (c-d) MRI modalities between the proposed HCA-Net (b and d) and the pose estimation method~\cite{azad2021stacked} (a and c). Green dots denote ground truth.} \label{fig:results}
\end{figure}

In \Cref{fig:results}, we provide a visual comparison between our HCA-Net and the pose estimation approach~\cite{azad2021stacked} in both T1w and T2w modalities. This comparison highlights the precision of our predictions. While the pose estimation approach misses one intervertebral location in T1w modality, our method successfully recognizes all intervertebral locations, with predictions closely matching the actual locations. This visual demonstration underscores the superior performance and accuracy of our HCA-Net.

Comparing our approach to the alternatives, we observe several key advantages. First, HCA-Net eliminates the need for complex preprocessing steps, such as image straightening or spinal cord region detection used in~\cite{rouhier2020spine}, making it more efficient and user-friendly. Second, our approach takes into account spatial relationships between IVDs, contributing to its superior performance, especially in FNR reduction.

\vspace{-0.5cm}
\section{Conclusion}
We proposed HCA-Net, a novel framework that capitalizes on a stack of hierarchical attention blocks to effectively encode both local and global information, ensuring precise localization of IVDs. The incorporation of a skeleton loss function further fine-tunes network predictions by considering the geometry of the intervertebral column. Through comprehensive experimentation, HCA-Net consistently demonstrated superior performance, attaining SOTA results.

\bibliographystyle{IEEEbib}
\bibliography{refs}

\end{document}